\pdfoutput=1
\documentclass{article}
\usepackage[letterpaper,top=2cm,bottom=2cm,left=3cm,right=3cm,marginparwidth=1.75cm]{geometry}

\usepackage[utf8]{inputenc} 
\usepackage[T1]{fontenc}    
\usepackage{hyperref}       
\usepackage{url}            
\usepackage{booktabs}      
\usepackage{amsfonts}      
\usepackage{nicefrac}       
\usepackage{microtype}      
\usepackage{xcolor}        
\usepackage{amsmath}
\usepackage{amssymb}
\usepackage[section]{placeins}
\usepackage{graphicx}
\usepackage{natbib}
\usepackage{booktabs}
\usepackage{xcolor}
\usepackage{tabularx}
\usepackage{hyperref}
\usepackage{stfloats}
\usepackage{algorithm,algorithmicx}
\usepackage[noend]{algpseudocode}
\usepackage{algcompatible}
\graphicspath{{./figures/}}
\usepackage{multirow}

\newcommand{\ModelName}{DetGPT}

\usepackage[skins]{tcolorbox} 
\tcbuselibrary{breakable} 
\newtcolorbox[auto counter, number within=section, list type=subsubsection, list inside=toc]{sectionbox}[2][]{
colback=white!98!gray, colframe=black, 
colbacktitle=white!90!gray, coltitle=black, 
fonttitle=\bfseries,
title={#2}, 
list entry={Comment \thetcbcounter\quad}
}

\usepackage{soul}

\title{DetGPT: Detect What You Need via Reasoning}

\author{\textbf{Renjie Pi}$^1$\thanks{Equal Contribution} \quad \textbf{Jiahui Gao}$^{2}$\footnotemark[1] \quad \textbf{Shizhe Diao}$^1$\footnotemark[1] \quad \textbf{Rui Pan}$^1$ \quad \and
\textbf{Hanze Dong}$^1$ \quad \textbf{Jipeng Zhang}$^1$ \quad
\textbf{Lewei Yao}$^1$\quad \textbf{Jianhua Han}$^3$\quad 
\textbf{Hang Xu}$^2$\quad  \and
\textbf{Lingpeng Kong}$^2$\quad
\textbf{Tong Zhang}$^1$ \quad
\\
  $^1$The Hong Kong University of Science and Technology\\
$^2$The University of Hong Kong\quad $^3$Shanghai Jiao Tong University \\
\texttt{\{rpi,sdiaoaa,rpan,hdongaj,jzhanggr,lyaoak\}@ust.hk,} \texttt{sumiler@hku.hk,} \\
\texttt{xbjxh@live.com},  \texttt{hanjianhua2012@gmail.com},\\
\texttt{lpk@cs.hku.hk},  \texttt{tongzhang@tongzhang-ml.org}
}

\begin{document}

\maketitle

\begin{abstract}
In recent years, the field of computer vision has seen significant advancements thanks to the development of large language models (LLMs). These models have enabled more effective and sophisticated interactions between humans and machines, paving the way for novel techniques that blur the lines between human and machine intelligence. In this paper, we introduce a new paradigm for object detection that we call \textbf{reasoning-based object detection}. Unlike conventional object detection methods that rely on specific object names, our approach enables users to interact with the system using natural language instructions, allowing for a higher level of interactivity. 
Our proposed method, called \textbf{{\ModelName}}, leverages state-of-the-art multi-modal models and open-vocabulary object detectors to perform reasoning within the context of the user's instructions and the visual scene. 
This enables {\ModelName} to automatically locate the object of interest based on the user's expressed desires, even if the object is not explicitly mentioned. 
For instance, if a user expresses a desire for a cold beverage, {\ModelName} can analyze the image, identify a fridge, and use its knowledge of typical fridge contents to locate the beverage. 
This flexibility makes our system applicable across a wide range of fields, from robotics and automation to autonomous driving. 
Overall, our proposed paradigm and {\ModelName} demonstrate the potential for more sophisticated and intuitive interactions between humans and machines. 
We hope that our proposed paradigm and approach will provide inspiration to the community and open the door to more interative and versatile object detection systems. Our project page is launched at \href{https://detgpt.github.io}{detgpt.github.io}.
\end{abstract}
\section{Introduction}
In recent years, the natural language processing field has seen remarkable advancements in the development of increasingly large language models (LLMs). 
LLMs such as GPT-3 {\citep{brown2020language}}, Bloom {\citep{scao2022bloom}}, PaLM {\citep{chowdhery2022palm}}, Megatron-Turing-530B {\citep{smith2022using}}, Chinchilla {\citep{hoffmann2022training}}, and others have expanded the horizons of language understanding and generation. These neural networks, with hundreds of billions of parameters, exhibit human-like proficiency in complex reasoning~\citep{wei2022chain, wang2022self, zhou2022least, zhang2022automatic, diao2023active, shum2023automatic}. 
However, the most powerful large models are closed-source {\citep{liang2022holistic}}, limiting their accessibility and hindering progress in the field. In response, Meta's LLaMA~\citep{touvron2023llama} offers a suite of powerful open-source models that bolster language model research. 
Recent works, such as Alpaca {\citep{alpaca}}, Vicuna {\citep{vicuna2023}}, and LMFlow {\citep{lmflow}}, have demonstrated impressive capabilities in instruction-following tasks and conversational applications after instruction tuning. Simultaneously, breakthroughs in the image and multimodal processing, as exemplified by models like LLAVA {\cite{liu2023llava}} and MiniGPT-4 {\cite{zhu2023minigpt4}}, have facilitated image-based interactions with robotic systems. These cutting-edge innovations are highly promising for a diverse array of applications across numerous fields.

The field of embodied AI / robotics is set to experience a significant transformation with the rise of powerful LLMs and multi-modal models. 
This is due to the heavy reliance on human-robot interactions, as highlighted by studies such as \citep{shah2023lm, brohan2022rt, fang2020graspnet}. 
With the advancement of LLMs and multi-modal models, robots will be able to interpret human instructions and reason over visual scenes, enabling them to execute corresponding actions. 
This breakthrough will lead to the creation of intelligent robots that are more helpful to humans, as they can effectively communicate with humans using natural language instructions, enhancing their utility and accessibility for a wider range of users. 
This exciting development has implications for various fields, including healthcare, manufacturing, and entertainment.

However, it is important to note that while multi-modal models have made remarkable progress in conversing with multi-modal inputs, this alone is insufficient for robots to interact with the physical world effectively. 
To achieve this, robots must be able to accurately identify objects within visual scenes, which is a vital prerequisite for performing actions such as "moving" and "grasping" objects. 
This goal of "localizing objects" is closely linked to the field of object detection, which is one of the most fundamental and extensively studied research areas in computer vision. 
Conventional object detection systems, such as Faster-RCNN~\citep{ren2015faster}, Retina-Net~\citep{lin2017focal}, and YOLO~\citep{redmon2016you}, have a fixed number of classification heads, which restricts practicality and confines predictions to only those classes that the model has been trained on. 
Recently, a series of open-vocabulary detection systems have emerged as the new trend~\citep{gu2021open,li2022grounded,yao2022detclip,liu2023grounding}.
Specifically, those models adopt the contrastive learning approach to align the object-level visual features with the textual class embeddings extracted from a pretrained text encoder (e.g., BERT~\citep{devlin-etal-2019-bert}). 
In this way, those models are able to correctly classify a much wider range of objects during inference.

Despite the success achieved by open-vocabulary object detection systems, they still require humans to provide accurate categories of the objects to be detected, which is neither user-friendly nor realistic due to the following reasons: 
(1) Humans are not always capable of supplying accurate object categories due to limited memory or knowledge. 
For instance, a user may want to locate fruits rich in vitamin K but may lack the specific knowledge of which fruits meet this criterion. 
In such cases, it would be advantageous for the model to autonomously reason about vitamin K-rich fruits and accurately detect and identify them. 
(2) The object categories that humans can provide are inherently limited and non-exhaustive. 
For instance, when detecting external behaviors that may pose an impact or danger to autonomous vehicles in driving, humans may only enumerate a few scenarios, such as restricted visibility, complex pedestrian traffic, and sudden lane changes by preceding vehicles. 
If the query "detect objects that are dangerous to autonomous vehicles" is directly assigned to the detection model, it can employ its own knowledge to identify a broader range of dangerous behaviors. 
Human knowledge is restricted, and the object categories that can be listed are likewise finite. 
In summary, we claim that large language models (such as ChatGPT) are promising in assisting in multimodal reasoning.
With powerful reasoning abilities, the instructions from humans can be simplified and the resulting answers will be more accurate and comprehensive.

In light of the above limitations of object detection systems, we propose a new research task: \textbf{reasoning-based object detection.} 
In essence, humans provide abstract queries via natural language, and the model discerns and reasons which object in the image may fulfill the query, subsequently detecting them. 
We made preliminary explorations in this direction. 
Specifically, we fine-tune multi-modal model built on LLMs to predict objects of interest based on user queries (instructions) and input images. 
We then provide the object names to an open-vocabulary detector for specific location prediction. 
This approach allows the model to analyze images and accurately predict the location of objects of interest based on user instructions. 
To facilitate the instruction-following ability of the model, we curate a high-quality fine-tuning dataset with 5000 images and around 30000 query-answer pairs, which is open-sourced for the research community.

\begin{figure}
\includegraphics[width=1.0\textwidth]{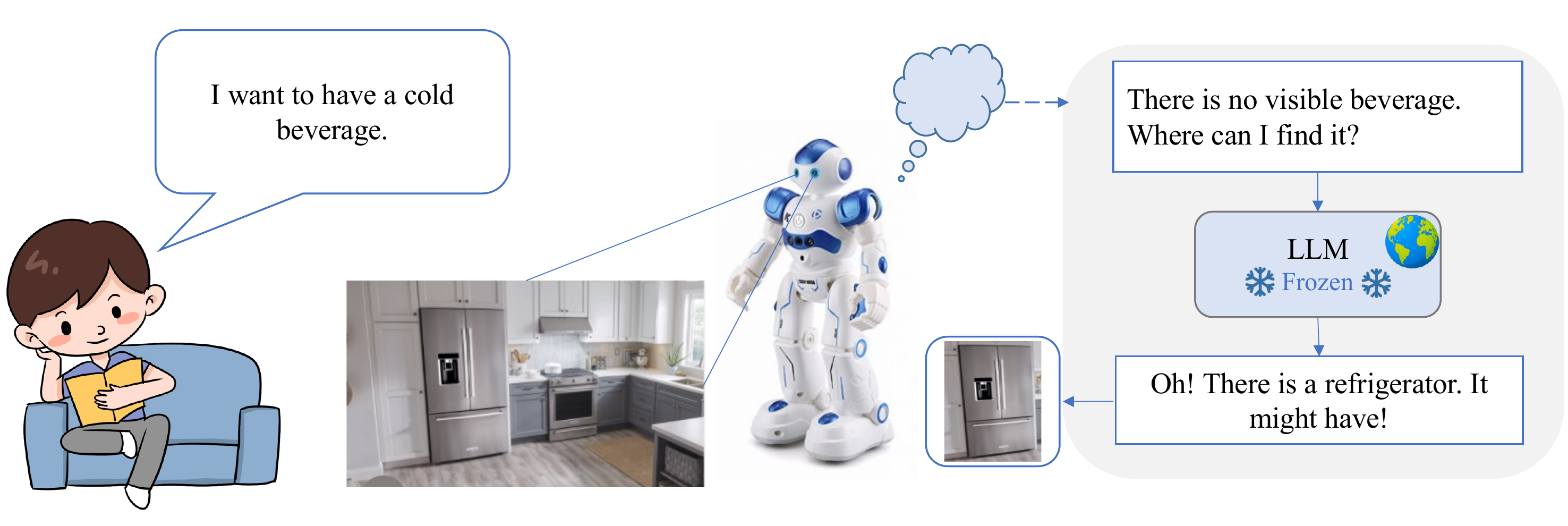} 
\caption{ The illustration of reasoning-based object detection task. The detection system is able to interpret human instruction, reason about the visual scene with common sense knowledge, and finally output the objects of interest. During the process, the LLM acts as the brain, while the detector empowers the system with the ability to "see". 
}\label{fig:framework}
\end{figure}
\section{Related Work}
\subsection{Large Language Models}

Recent years have seen significant progress in developing increasingly large and powerful language models.
Models like GPT-3~\citep{brown2020language}, Bloom~\citep{scao2022bloom}, PaLM~\citep{chowdhery2022palm}, megatron-turing-530b~\citep{smith2022using}, Chinchilla~\citep{hoffmann2022training} and so on have pushed the capabilities of language understanding and generation to new frontiers.
These massive neural networks, with hundreds of billions of parameters, have demonstrated human-level abilities in text classification, text generation, knowledge-intensive tasks, and even complex reasoning tasks.
However, most of the powerful models are deployed in the cloud by commercial companies.
For example, according to a holistic study~\citep{liang2022holistic}, closed-source models (e.g., InstructGPT~\citep{ouyang2022training} and Anthropic-LM~\citep{bai2022training}) usually outperforms open-source models (e.g., GPT-J~\citep{gpt-j} and OPT~\citep{zhang2022opt}), posting a challenge to open-source research. 
Recently, Meta's LLaMA~\citep{touvron2023llama} provides a series of powerful open-source models that boost language model research.
For example, recent Alpaca~\citep{alpaca}, Vicuna~\citep{vicuna2023}, and LMFlow~\citep{lmflow} have demonstrated the powerful capacity of using LLaMA as a base model, with promising performance on instruction-following tasks and chitchat after instruction tuning.

\subsection{Large Multi-Modal Models}
Given the success of language models, many following research explored vision-language interaction, resulting in a number of multi-modal models.
The development in multi-modal learning displays a clear path following the language model research.
Inspired by BERT-like encoder models, most of the multi-modal models~\citep{lu2019vilbert,tan2019lxmert,chen2020uniter,li2021align,li2020oscar} before 2021 are encoder-only Transformers, which are good at cross-modal understanding tasks.
However, the transition from encoder-only models to decoder-based models in language model research inspires the pattern shift in multi-modal learning, including encoder-decoder models like VL-T5~\citep{cho2021unifying}, OFA~\citep{wang2022unifying}, DaVinci~\citep{diao2023write} and decoder-only models like GPT-4~\citep{openai2023gpt4}.
Most recently, we have witnessed the potential of multimodal interaction due to the powerful language abilities of LLaMA. 
Relevant works include LLaVA~\citep{liu2023llava} and minigpt4~\citep{zhu2023minigpt4}. 
Unlike these works, our research incorporates more powerful object detection capabilities, allowing for greater control over objects through language.

\subsection{Object Detection}
Object detection is one of the most fundamental tasks in computer vision, which aims at localizing objects in images. Traditional object detectors have a fixed number of classification heads, which restricts their practicality and makes them only capable of predicting the classes on which they are trained~\citep{girshick2015fast, ren2015faster, lin2017focal, yao2021g, duan2019centernet, yao2021joint, zhu2020deformable, carion2020end}. Recently, open-vocabulary object detection has attracted a lot of attention~\citep{gu2021open, li2022grounded, liu2023grounding, yao2022detclip}. The main philosophy is to utilize contrastive training between object visual features and text phrases. In such a way, object detectors are able to recognize objects that are unseen during training based on their semantics. Despite the success of open vocabulary object detectors, their ability is still limited in the sense that they can only perform prediction given specific object phrases. On the other hand, our DetGPT not only enables localizing objects given high-level user instructs but also empowers the detector with reasoning capability via the knowledge stored in the large language models.
\section{Problem Statement}
Recent multi-modal models combined with LLMs have shown promising results in visual understanding and reasoning based on the visual scene and natural language input. 
However, they still lack the ability for fine-grained visual understanding and precise localization, which makes them difficult to be applied to scenarios involving embodied AI, such as robotics ad autonomous driving. 
On the other hand, object detection is a crucial task in computer vision, which enables models to analyze images in a fine-grained manner and predict precise locations for visible objects. 
Unfortunately, existing closed-set and open-vocabulary detection methods rely on specific category phrases that may not accurately describe the objects in the image.

To address these limitations, we propose a new task termed \textbf{reasoning-based object detection}. In this task, users provide abstract queries using natural language, and the model analyzes both the image and user input, reasons about which objects in the image may fulfill the user's goal, and finally detects their location in the image. 
For example, as shown in Figure \ref{fig:framework}, when a user requests "I want to have a cold beverage," the model first analyzes the image of a kitchen and determines that there is no "cold beverage" available. 
Then, it identifies a refrigerator in the image and, based on the common sense knowledge stored in the LLM, infers that the refrigerator may store a cold beverage.

The proposed task of reasoning-based object detection opens up a world of possibilities for human-machine interactions. 
For instance, it allows users to ask the robot to detect missing ingredients in the fridge while cooking or locate lost keys in the house. 
They can also ask the robot to detect items that need to be cleaned or find a specific book in their library. 
These examples demonstrate how reasoning-based object detection can greatly improve the capabilities of domestic robots and make human-robot interactions more intuitive and natural. 
This approach can also lead to the development of more advanced and user-friendly applications in various domains, such as home automation, healthcare, and education.

\section{Query-Answer Instruction Data Generation}
The traditional way to label a dataset for instruction tuning requires a large amount of human labor. 
Recently, large language models (LLMs) such as ChatGPT have been shown to possess superior generation capability, which can be used to replace human labeling with automatically generated annotations~\citep{SchickS21a,ye2022zerogen,ye-etal-2022-progen,meng2022generating,gao2023selfguided}. 
However, the difficulty that lies in data generation with image inputs is that text-only LLMs are not able to interpret images, which poses challenges for them to generate annotations that are related to visual inputs.

Motivated by LLAVA~\cite{liu2023llava}, we leverage the pre-existing datasets for image captioning and object detection, and employ two types of textual annotations to bridge the gap between visual and textual representations: (1) Image Captions, which offer depictions of the visual content from different viewpoints. (2) Objects categories, which are the objects present in the image.
Based on the given captions and objects, we design query-answer prompts to instruct ChatGPT to generate the following: (1) a more detailed description of the scene. 
This will give ChatGPT itself a better sense of the visual scene, which facilitates the generation of instruction annotations; 
(2) query-answer pairs, which consist of a user instruction (query) and the corresponding answer that contains both reasoning and items' names in the image that match the instruction. 
For each image, we generate one detailed description followed by several instruction-answer pairs. 
We then reorganize the annotations such that each image is associated with the corresponding instruction-answer pairs. 
The detailed system prompt for our cross-modal object detection task is shown in Table~\ref{tab:prompts}. 
To enable better annotation generation, we further manually design two in-context examples for querying ChatGPT, which are shown in Appendix~\ref{tab:in-context examples}. An example of the generated query-answer pairs is shown in Table ~\ref{tab:full_example_car_bbox}. 

\section{Method}
\subsection{Model Architecture}
\begin{figure}[h]
\centering
\includegraphics[width=\textwidth]{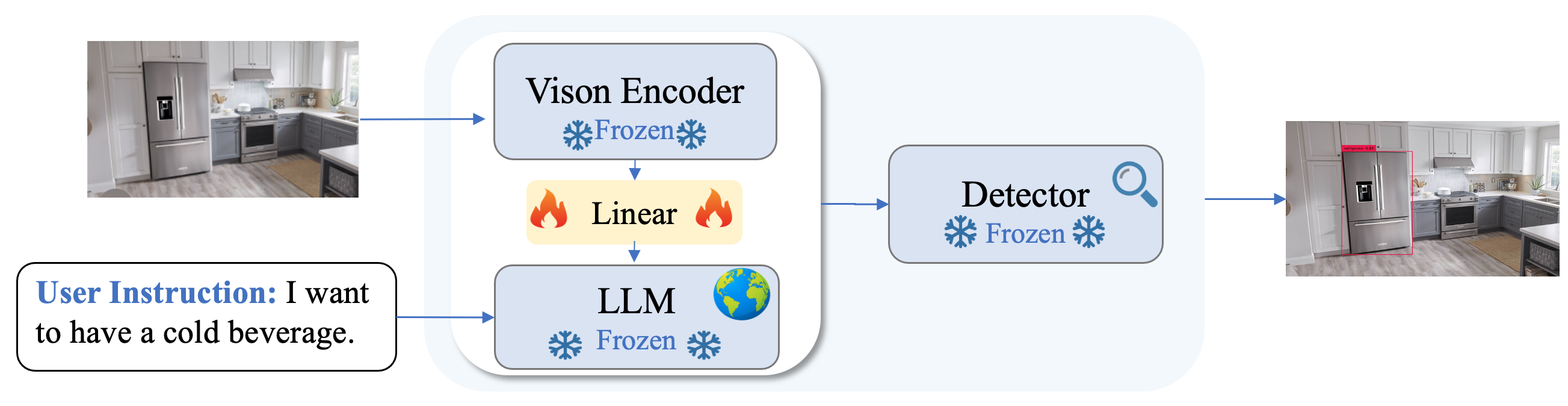} 
\caption{Framework of DetGPT. The multi-model model consisted of vision encoder and LLM interprets the user instruction, reasons over the visual scene, and finds objects matching the user instruction. Then, the object names/phrases are passed to the open-vocabulary detector for localization.
}\label{fig:model_arch}
\end{figure}

As an initial attempt towards reasoning-based object detection, we propose a two-stage approach that first leverages the multi-modal model to interpret the image and list out the relevant objects names/phrases that match the user's instructions; 
then we leverage an open-vocabulary object detector to localize the relevant objects given the results from the multi-modal model. 
Specifically, for the multi-modal model, we employ a pre-trained visual encoder to extract image features, followed by a cross-modal alignment function to map the image features to the text domain. Then, we utilize a pre-trained LLM as the knowledge brain to interpret both the image features and human instructions, perform reasoning, and determine the target objects in the image that can help fulfill the given user query. 
Our framework is illustrated in Figure~\ref{fig:model_arch}.

\paragraph{Implementation Details.} Inspired by\citep{zhu2023minigpt4}, we employ the visual encoder of BLIP-2~\cite{li2023blip2} as the vision encoder and utilize 13B Vicuna~\cite{vicuna2023} as the language model to interpret both visual and text features. For the open-vocabulary detector, we leverage Grounding-DINO~\cite{liu2023grounding} to localize the target objects in the image.
Following MiniGPT-4, we employ a linear projection layer for the cross-modal alignment, which has been proven effective in bridging the gap between vision and language modalities.

\subsection{Training}
The main challenge of our framework is to enable the multi-modal model to name all the related object categories in the image that match the user's query. This requires the model to understand the image and user instructions, performing reasoning, and then finally retrieve the objects of interest. To overcome this challenge, we propose the following steps that can be executed in sequence: 

\textbf{Step 1. Image-text Pretraining.} We follow \cite{zhu2023minigpt4} and leverage a combined dataset of SBU, LAION and Conceptual Caption to conduct image-text pretraining for visual-textual feature alignment. During the process, only the projection linear layer is trainable, while all other components are kept frozen. Since the visual feature from BLIP-2~\cite{li2023blip2} is already well aligned with textual features, tuning just the linear layer already suffices for aligning with a new language model such as Vicuna. 

\textbf{Step 2. Instruction Tuning.} After the first step, although the model is able to generate detailed descriptions based on the image, it is not able to well interpret human instruction and list out objects related to the user's goal. Therefore, we leverage our curated query-answer instruction dataset to fine-tune our model. Similar to the first step, only the linear projection layer is made tunable. During the training process, the model receives the image and the instruction as inputs, and is trained to generate the corresponding answer by minimizing the following language modeling loss:
\begin{align}
    \mathcal{L}=-\sum^L_{t=1}\log p\left[y^{i,j}_t | \mathcal{F} (y^{i,j}_{(<t)}, I^i)\right]
\end{align}
where $\mathcal{F}$ represents the multi-modal model. $I^i$ represents the $i^{th}$ image, and $y^{i,j}_t$ denotes the $t^{th}$ token of the $j^{th}$ answer that belongs to the $i^{th}$ image. $L$ is the length of the answer. We demonstrate that the instruction tuning phase empowers the language model to comprehend fine-grained image features and summarize the matching objects' categories in a specific pattern after reasoning.

Even though instruction tuning already empowers the model to identify the objects of interest in the image, we find that the output format of the model often varies, which poses difficulty in extracting the relevant object names/phrases. Therefore, as shown in Table \ref{tab:user_prompts}, we design the user prompt that is helpful for the model to output the objects strictly in a given format, which makes our model more stable. The final input sequence used to train the model is "\#\#\#Human: $\langle$ Img $\rangle$ $\langle$ \textcolor{blue}{ImageHere} $\rangle$ $\langle$ Img$ \rangle$ $\langle$ \textcolor{red}{TextHere} $\rangle$ $\langle$ User\_Prompt $\rangle$". \textcolor{blue}{Blue color} represents the input image and \textcolor{red}{Red color} represents user instruction. 

\begin{table*}[h]\centering
\begin{minipage}{1.0\columnwidth}\vspace{0mm}
\centering
\begin{sectionbox}[]{User Prompt} 
    \centering
   
      \footnotesize
    \begin{tabular}{p{0.97\columnwidth} c}
Answer me with several sentences. End the answer by listing out target objects to my question strictly as follows: $\langle$Therefore the answer is: [object\_names]$\rangle$.
    \end{tabular}
\end{sectionbox}
\vspace{-2mm}
\caption{User Prompt. 
We found that prompting is necessary for listing names of objects of interest in a consistent format, which makes DetGPT more stable.
}
    \label{tab:user_prompts}
\end{minipage}
\end{table*}

\textbf{Step 3. Inference.} During inference, we first provide the model with a system prompt (show in Table \ref{tab:sys_prompts}), which we find to be helpful in stablizing the model's output. In addition, we append the user prompt after the user's query similar to instruction tuning. After obtaining the generated answer from the multi-modal model, we extract the object names/phrases from it by matching the specific output format, .i.e, the object names following "Therefore the answer is: ". Finally, we send the names/phrases and the image to the object detector for localization.

\begin{table*}[h!]\centering
\begin{minipage}{1.0\columnwidth}\vspace{0mm}    \centering
\begin{sectionbox}[]{System Prompt for DetGPT inference}
    \centering
      \footnotesize
    \begin{tabular}{p{0.97\columnwidth} c}
You must strictly answer the question step by step:\\
Step-1. describe the given image in detail.\\
Step-2. find all the objects related to user input, and concisely explain why these objects meet the requirement.\\
Step-3. list out all related objects existing in the image strictly as follows: $\langle$ Therefore the answer is: [object\_names] $\rangle$.\\
Complete all 3 steps as detailed as possible.\\
You must finish the answer with a complete sentence.\\
    \end{tabular}
\end{sectionbox}
\vspace{-2mm}
\caption{System Prompt provided to DetGPT during inference. We found that prompting is necessary for listing names of object of interest in a consistent format, which makes DetGPT more stable.}
    \label{tab:sys_prompts}
\end{minipage}
\end{table*}

\vspace{-5mm}
\begin{table*}[t!]\centering
\begin{minipage}{1.0\columnwidth}\vspace{0mm}    \centering
\begin{sectionbox}[]{Generated Examples}
    \centering
   
      \footnotesize
    \begin{tabular}{p{0.97\columnwidth} c}
{ {\bf Captions:} } &\\
Some people in a grass field flying a kite in the sky. & \hspace{-3.2cm} \multirow{5}{*}{ \includegraphics[height=2.0cm]{000000014439.jpg} }
\\
A boy playing with a kite in the park.\\
a colorful kite flying by some people in the park.\\
A woman and a young girl holding a kite on a green field.\\
A group of people standing on a field flying a colorful kite.
\\
\\
{\bf Objects: }  & \\
chair, kite, backpack, person\\
\hrulefill & \\
\textbf{Description:}\\
The image shows a group of people on a grassy field, flying a colorful kite in the sky. Some of the people are standing while others are sitting on chairs. One woman and a young girl are holding onto the kite while it flies in the air. There are trees and other greenery visible in the background.\\
\\
\textbf{Query and Answer:}\\  
1. Query: I want to fly a kite. What object do I need? Answer: In the image, there is a kite present, which you can use to fly a kite. Therefore the answer is: [kite]\\
2. Query: Find all the people in the image. Answer: In the image, there are multiple people visible, standing and sitting, while flying a kite. Therefore the answer is: [person]\\
3. Query: I want to sit while flying a kite. What object can I use? Answer: In the image, there are chairs present on the grassy field, which you can use to sit while flying a kite. Therefore the answer is: [chair]\\
4. Query: Find all the objects that are colorful. Answer: In the image, there are multiple objects visible, including the kite flying in the sky, which is colorful. Therefore the answer is: [kite]\\
5. Query: Find all the objects that are used for carrying things. Answer: In the image, there is a backpack visible, which is used for carrying things. Therefore the answer is: [backpack]\\
\rule{\linewidth}{1pt} & \\
{ {\bf Captions:} } &\\
Two computers are sitting on top of the desk. & \hspace{-3.2cm} \multirow{5}{*}{ \includegraphics[height=2.0cm]{000000340894.jpg} }
\\
Two computers on a large wooden computer desk.\\
A desk with a keyboard, laptop and monitor.\\
A computer monitor next to a keyboard, laptop and a mouse.\\
A laptop with external keyboard, mouse, phone and photo on a desk.\\
\\
{\bf Objects: }  & \\
chair, person, tv, cell phone, cup, laptop, mouse, keyboard\\
\hrulefill & \\
\textbf{Description:}\\
The image displays a desk with various computer accessories on it. There are two computers present on the desk, which appear to have external keyboards attached to them. In addition to the computers, there is a laptop, computer monitor, and mouse present on the desk. A phone is visible as well, along with a photo frame. The desk appears to be made of wood, and there are no other visible objects or people in the scene.\\
\\
\textbf{Query and Answer:}\\  
1. Query: How can I make a phone call? Answer: In the image, there are keyboard, person, chair, laptop, mouse, cup, cell phone, and TV. To make a phone call, use the cell phone on the desk. Therefore the answer is: [cell phone]\\
2. Query: Find all the electronic devices in the image. Answer: In the image, there are keyboard, laptop, mouse, computer monitor, cell phone, and TV. All of these are electronic devices. Therefore the Answer: is: [keyboard, laptop, mouse, computer monitor, cell phone, TV]\\
3. Query: Find all the objects that can be used for typing. Answer: In the image, there are keyboard, laptop, and external keyboard. All of these can be used for typing. Therefore the answer is: [keyboard, laptop, external keyboard]\\
4. Query: Can you find any object that is black in color? Answer: In the image, there are keyboard, person, chair, laptop, mouse, cup, cell phone, and TV. The external keyboard and computer mouse are black in color. Therefore the answer is: [external keyboard, mouse]\\
5. Query: Detect all the objects that are rectangular in shape. Answer: In the image, there are keyboard, laptop, mouse, computer monitor, cell phone, and TV. Among them, the computer monitor and the laptop can be considered rectangular in shape. Therefore the answer is: [computer monitor, laptop]\\
\end{tabular}
\end{sectionbox}
\vspace{-2mm}
\caption{
Two examples to demonstrate the instruction-following data. The top section displays the prompts used to instruct text-only ChatGPT, which consists of captions and objects of the visual image. The bottom section shows the responses generated by ChatGPT.}
    \label{tab:full_example_car_bbox}
\end{minipage}
\end{table*}

\begin{table*}[t!]\centering
\begin{minipage}{1.0\columnwidth}\vspace{0mm}    \centering
\begin{sectionbox}[]{System Prompt for Data Generation}
    \centering
      \footnotesize
    \begin{tabular}{p{0.97\columnwidth} c}
You are an AI visual assistant, and you are seeing a single image. What you see are provided with a few sentences, which describe the same image you are looking at. Also, some objects that appear in th image are given to you. You are responsible for the following: 
first, you need to describe the image contents with necessary but not redundant details. 
then, you need to design queries that can be completed with objects that are shown in the image. \\
\\
You should generate query-answeer pairs that are goal-orienated, where the user inputs something he/she wishes to achieve, and you are responsible to find the objects in the image that helps he/she to do so. You should generate as many queries as possible, but the related objects must be contained in the five sentences or the given objects. Also, the queries for an image should be diverse, spanning across all types of queries mentioned above. \\
\\
Note that you should design attribute-related queries (such as color or shape), only when you are certain about it. Do not generate such queries if the captions provided to you does not contain such information. \\
\\
You must response any queries or answer in the following way:\\
Query:  $\langle$QUERY$\rangle$ Answer: $\langle$ANSWER$\rangle$ Therefore the answer is: $\langle$TARGET\_OBJETCTS$\rangle$\\
\\
The objects in $\langle$TARGET\_OBJETCTS$\rangle$ must be shown in the image, and can be used to solve the query in $\langle$QUERY$\rangle$.\\
\\
When answer each query, you must (1) describe all the object (you may refer to the complete object list), (2) based on common sense, use correct object(s) to answer the question, (3) list out target objects in the following manner: "Therefore the answer is: $\langle$TARGET\_OBJETCTS$\rangle$".
\\
\\
You should answer the query based on the your understanding visual feature.\\ 
    \end{tabular}
\end{sectionbox}
\vspace{-2mm}
\caption{System Prompt provided to ChatGPT to generate cross-modal instruction data for reasoning object detection task.}
    \label{tab:prompts}
\end{minipage}
\end{table*}
\section{Demonstration}
\begin{figure}
\centering
\includegraphics[width=0.9\textwidth]{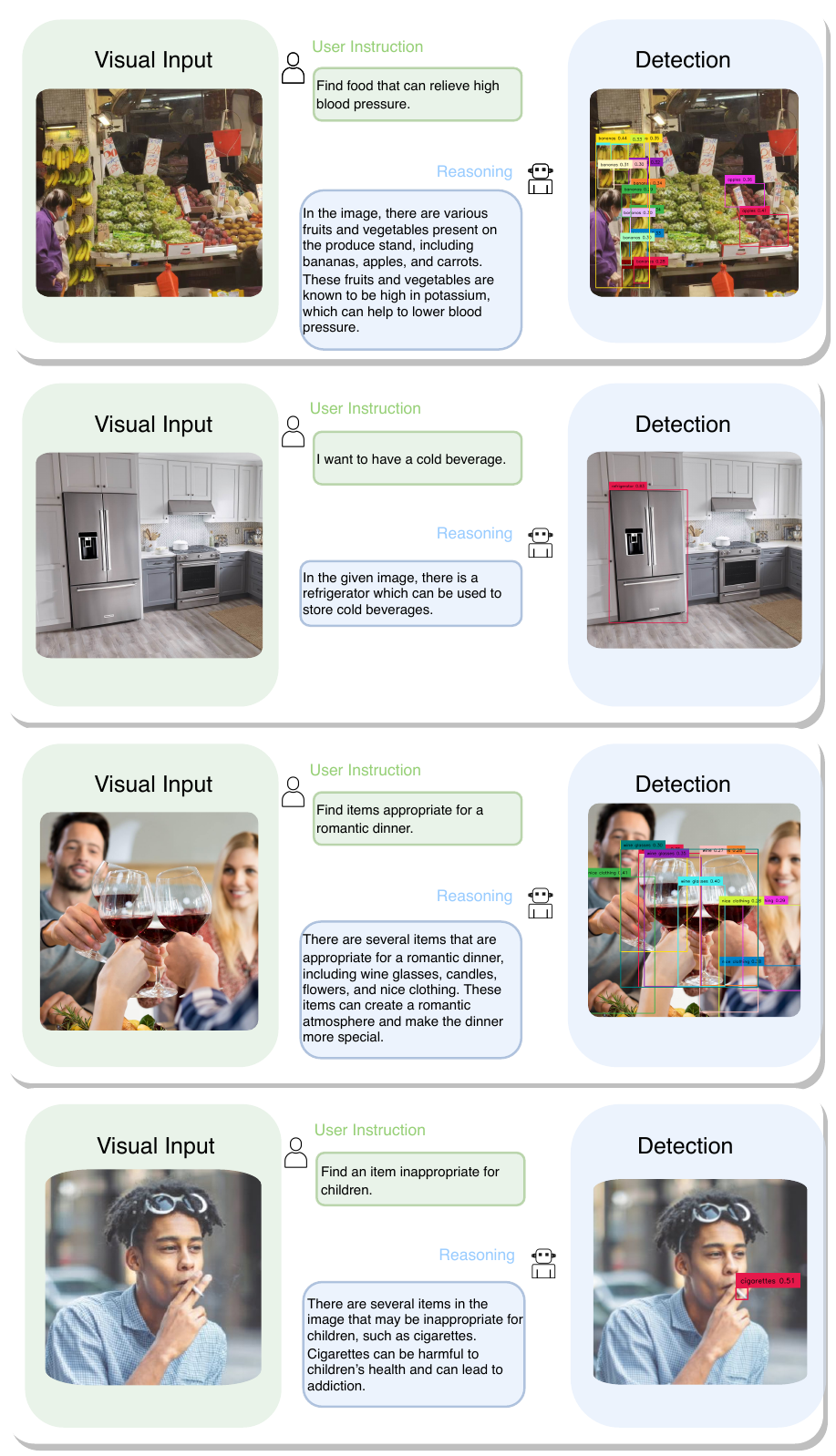} 
\caption{ Demonstration of the reasoning process and generated bounding boxes of our DetGPT. 
}\label{fig:demo1}
\end{figure}

\begin{figure}
\centering
\includegraphics[width=0.9\textwidth]{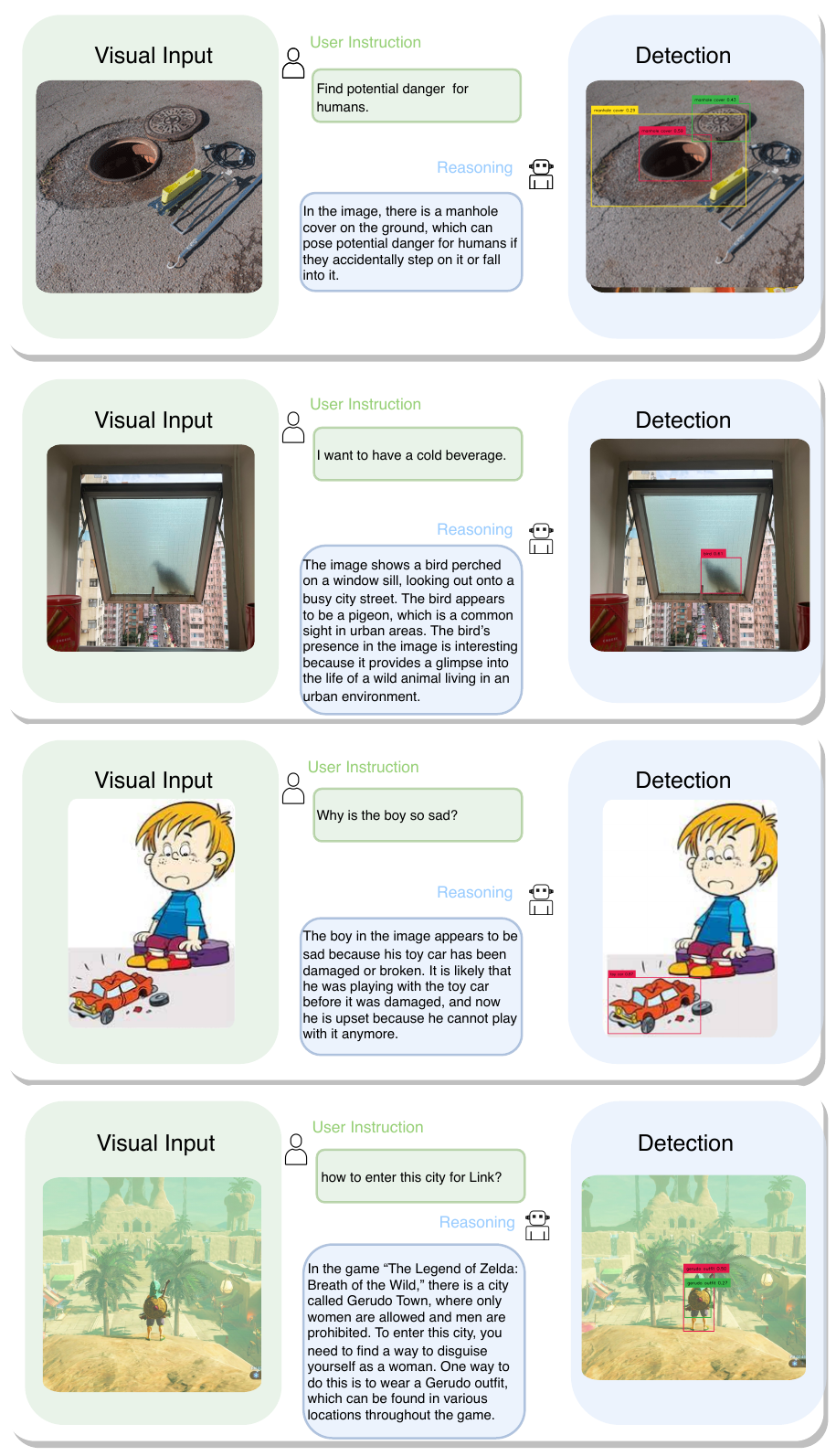} 
\caption{ Demonstration of the reasoning process and generated bounding boxes of our DetGPT. 
}\label{fig:demo2}
\end{figure}

This section presents the results of DetGPT in Figure \ref{fig:demo1} and Figure \ref{fig:demo2} and evaluates its capabilities. The rendered bounding boxes and reasoning process are both displayed. Interestingly, DetGPT exhibits several appealing features as follows:

Firstly, it is proficient in conducting common-sense reasoning based on the user's query and the image. For example, if the user requests a cold beverage, and the input image is a picture of a kitchen where no beverage is visible, DetGPT can reason over the visual scene and determine that a refrigerator may contain a beverage. It can then localize the fridge.

Secondly, DetGPT can utilize the rich knowledge stored in LLMs to conduct reasoning beyond human common sense. For instance, if the user wishes to identify the food that relieves high blood pressure, DetGPT can use its internal knowledge to determine that the lack of potassium is one of the causes of high blood pressure, which is present in bananas and apples.

Thirdly, it can locate all relevant objects based on the user's queries, even when the instructions are abstract and the related objects are non-exhaustive. For example, the instruction "the items that are inappropriate for children" encompasses a wide range of things, but DetGPT can identify the relevant object based on the image and subsequently detect the cigarette.

Finally, DetGPT can generalize to a broad range of objects that do not appear in the task tuning set. For instance, it can analyze a screenshot from a videogame and locate the game props required to pass the game level. Notably, our task-tuning dataset is constructed based on the COCO dataset, which only includes 80 daily object categories. This verifies that while task tuning provides DetGPT with the ability to identify the object of interest, the pretraining stage is crucial for its exceptional generalization ability.
\section{Limitation}
\begin{figure}
\centering
\includegraphics[width=0.9\textwidth]{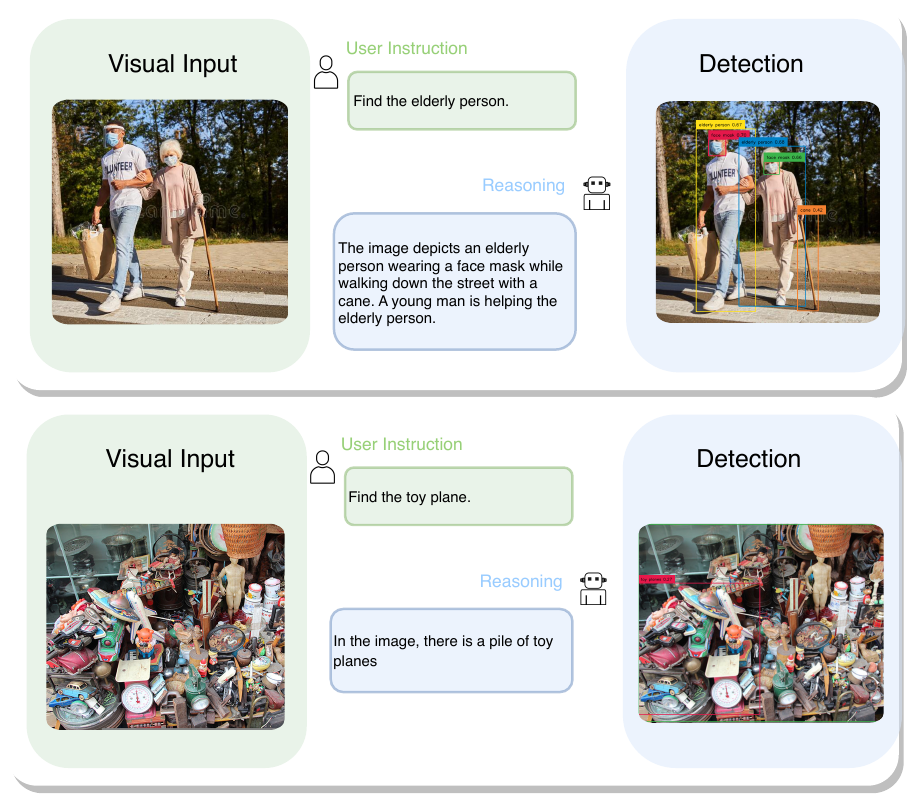} 
\caption{ Demonstration of failure cases. Top: even though the multi-modal is able to understand the visual scene and find the elderly lady, the object detector localizes both the young man and the elderly person, and label them both as elderly person. This may be due to the detector is not able to distinguish "young man" from "elderly person". Bottom: there is only one toy plane in the image, but the multi-modal model recognizes "a pile of toy planes". This may be caused by the multi-modal model's lack of fine-grained visual recognition capability.
}\label{fig:failure}
\end{figure}

As the first attempt towards a reasoning-based object detection system, despite the promising results, {\ModelName} still has some limitations (shown in Figure \ref{fig:failure}). 
Due to the two-stage nature of {\ModelName}, the weaknesses of both open-vocabulary detector and multi-modal models become the bottleneck. For example, we observe that in some cases, even though the multi-modal model is able to find the relevant objects from the image, the open-vocabulary detector is not able to localize them, which may be because the training data of the object detector does not encompass such visual concepts. In some other cases, the multi-modal model is not able to find all relevant objects in the image, possibly due to the lack of fine-grained visual recognition ability.
The above limitations promote new research in this direction and demand more advanced solutions.

\section{Conclusion}
In this paper, we propose a new task termed \textbf{reasoning-based object detection}, which requires the model to interpret human instructions, reason over the visual scene, and finally localize the objects of interest. 
As an initial attempt towards this task, we design a two-stage detection pipeline, which first uses a multi-modal model to derive the objects that match the user's query from in the image, then use an off-the-shelf open-vocabulary object detector to localize those objects in the image. 
To empower the model with the ability to find relevant objects in the image, we construct a dataset for task tuning with the help of ChatGPT.
We demonstrate that the resulting model is able to interpret human instruction and localize relevant objects, even if the objects are missing in the task-tuning dataset. 
We hope that our method will pave the way for a more interactive and user-friendly object detection system, which will inspire later works on embodied AI, autonomous driving, and robotics.


\appendix
\section{In-context Examples}\label{tab:in-context examples}
\begin{table*}[t!]\centering
\begin{minipage}{1.0\columnwidth}\vspace{0mm}    \centering
\begin{sectionbox}[]{In Context Example 1}
    \centering
   
      \footnotesize
    \begin{tabular}{p{0.97\columnwidth} c}
{ {\bf Captions:} } &\\
Three computer screens sitting on a desk with headphones attached.\\
A desk that has a few computer screens on it.\\
A desk with a computer, a laptop, headphones and a telephone office desk with a laptop , headphones, keyboard and dual monitors.\\
A home computer and keyboard on a desk.\\
\\
{\bf Objects: }  & \\
computer, monitor, keyboard, book, mouse, headphone, telephone \\
\\
\textbf{Description:}\\
The image depicts a desk with multiple computer screens and various accessories placed on it. Specifically, there are three computer screens present on the desk, each of which has headphones attached to them.  In addition to the screens, there is also a laptop, a keyboard, and a telephone present on the desk.  The headphones appear to be over-ear style and are connected to the screens via cords. The keyboard is positioned in front of the laptop, and the telephone is located to the left of the screens.\\
\\

\textbf{Query and Answer:}\\
1.Query: I wish to watch a movie. What can I use? Answer: In the image, there are computer, monitor, keyboard, book, mouse, headphone and telephone. You can use computers and laptops to watch movies online. Therefore the answer is: [computer, monitor]\\
2.Query: It is late, and I wish to read before going to bed. Answer: In the image,  there are computer, monitor, keyboard, book, mouse, headphone and telephone.  If you wish to read, you could either read from online sources using the computer, or read from the book. Therefore the answer is: [computer, monitor, book]\\
3.Query: Find all the computer, mouse, keyboard and cake in the image. Answer: In the image, there are computer, mouse and keyboard, but there is no presence of cake. Therefore the answer is: [computer, mouse, keyboard]\\
4.Query: I need to call someone. What object can I use in the image? Answer: In the image, there are computer, monitor, keyboard, book, mouse, headphone and telephone. You can use the telephone on the desk to make phone calls. Therefore the answer is: [telephone]\\
5.Query: What can I use to listen to music? Answer: In the image, there are computer, monitor, keyboard, book, mouse, headphone and telephone. You can use the headphone on the desk to listen to music. Therefore the answer is: [headphone]\\
6.Query: Detect airplanes and computer in the image? Answer: In the image, there there are computer, monitor, keyboard, book, mouse, headphone and telephone. Therefore, I can detect the computer in the image. However, no airplane can be found. Therefore the answer is: [computer]\\
7.Query: Detect people and telephone in the image? Answer: In the image, there are computer, monitor, keyboard, book, mouse, headphone and telephone. However, no people can be found. Therefore the answer is: [telephone]\\
8.Query: Is there any object that is rectangular in shape. Answer: In the image,there are computer, monitor, keyboard, book, mouse, headphone and telephone. The monitors and the laptop are rectangular in shape. Therefore the answer is: [monitor, laptop]\\
9.Query: Detect computer in the image? Answer: In the image, there are computer, monitor, keyboard, book, mouse, headphone and telephone. Therefore the answer is: [computer]\\
10.Query: Detect cake and computer in the image? Answer: In the image, there are computer, monitor, keyboard, book, mouse, headphone and telephone. However, no cake can be found. Therefore the answer is: [computer]\\

    \end{tabular}
\end{sectionbox}
\vspace{-2mm}
\caption{In-context example.}
    \label{tab:full_example_car_bbox}
\end{minipage}
\end{table*}

\begin{table*}[t!]\centering
\begin{minipage}{1.0\columnwidth}\vspace{0mm}    \centering
\begin{sectionbox}[]{In Context Example 2}
    \centering
   
      \footnotesize
    \begin{tabular}{p{0.97\columnwidth} c}
{ {\bf Captions:} } &\\
A group of people standing outside of a black vehicle with various luggage. 
\\
A person holding a tennis racket about to serve a ball on a tennis court. \\
A group of people playing tennis on a sunny day. \\
A tennis player about to serve the ball on a clay court. \\
Two people playing tennis on a green court with a blue sky in the background. \\
\\
{\bf Objects: }  & \\
tennis racket, tennis ball, tennis court, net, tennis shoes, water abottle\\
\\
\textbf{Description:}\\
The image shows a tennis court, with a net dividing it in half. A person is standing on one side of the court, holding a tennis racket and getting ready to serve a ball. The court surface appears to be clay, and there are green trees visible in the background. The person is wearing tennis shoes and has a water bottle next to them.\\
\\
\textbf{Query and Answer:}\\
1.Query: I want to practice my forehand. What object can I use? Answer: In the image, there are tennis racket, tennis ball, tennis court, net, tennis shoes and water abottle. You can use the tennis racket and tennis ball to practice your forehand. Therefore the answer is: [tennis racket, tennis ball] \\
2.Query: What can I use to hydrate myself during the game? Answer: In the image, there are tennis racket, tennis ball, tennis court, net, tennis shoes and water abottle. Using the water bottle present, you can hydrate yourself. Therefore the answer is: [water bottle]\\
3.Query: Find all the objects that are used in a tennis game. Answer: In the image, there are tennis racket, tennis ball, tennis court, net, and tennis shoes, all of which are used in a tennis game. Therefore the answer is: [tennis racket, tennis ball, tennis court, net, tennis shoes]\\
4.Query: I want to play tennis with my friend. Answer: In the image, there is a tennis court, tennis racket, and tennis ball present, which you can use to play tennis with your friend. Therefore the answer is: [tennis court, tennis racket, tennis ball] \\
6.Query: Find all the objects that are round in shape. Answer: In the image, there are tennis racket, tennis ball, tennis court, net, tennis shoes and water abottle. Among them, the tennis ball is round in shape. Therefore the answer is: [tennis ball] \\
7.Query: Find the ice cream and the tennis racket. Answer: In the image, there are tennis racket, tennis ball, tennis court, net, tennis shoes and water abottle.  However, there is no ice cream in the image. Therefore the answer is: [tennis racket]\\

    \end{tabular}
\end{sectionbox}
\vspace{-2mm}
\caption{In-context example.}
    \label{tab:full_example_car_bbox}
\end{minipage}
\end{table*}

\newpage
{\small
\bibliographystyle{plainnat}
\bibliography{custom}
}

\end{document}